\documentclass[journal]{IEEEtran}
\ifCLASSINFOpdf
\else
\fi

\usepackage{textcomp}
\usepackage{color}
\usepackage{soul}
\usepackage{graphicx}
\usepackage{pbox}
\usepackage{url}
\usepackage{siunitx}
\usepackage{amssymb}
\usepackage{ragged2e}
\usepackage{multirow}
\usepackage{hhline}
\usepackage{inputenc}

\begin{document}
%
\title{A Novel Multimodal Approach for Studying the Dynamics of Curiosity in Small Group Learning}
%
%
%

\author{Tanmay Sinha$^1$,\IEEEmembership{}
        Zhen Bai$^2$,\IEEEmembership{}
        Justine Cassell$^3$~\IEEEmembership{}
\thanks{$^1$T. Sinha is with Professorship for Learning Sciences and Higher Education, ETH Z{\"u}rich, Switzerland. E-mail: tanmay.sinha@gess.ethz.ch}
\thanks{$^2$Z. Bai is with Department of Computer Science, University of Rochester, USA. E-mail: zbai@cs.rochester.edu}
\thanks{$^3$J. Cassell is with School of Computer Science, Carnegie Mellon University, USA and Paris Artificial Intelligence Research Institute, Inria, France. E-mail: justine@cs.cmu.edu}}

\maketitle

\begin{abstract}
Curiosity is a vital metacognitive skill in educational contexts, leading to creativity, and a love of learning. And while many school systems increasingly undercut curiosity by teaching to the test, teachers are increasingly interested in how to evoke curiosity in their students to prepare them for a world in which lifelong learning and reskilling will be more and more important. One aspect of curiosity that has received little attention, however, is the role of peers in eliciting curiosity. We present what we believe to be the first theoretical framework that articulates an integrated socio-cognitive account of curiosity that ties observable behaviors in peers to underlying curiosity states. We make a bipartite distinction between individual and interpersonal functions that contribute to curiosity, and multimodal behaviors that fulfill these functions. We validate the proposed framework by leveraging a longitudinal latent variable modeling approach. Findings confirm a positive predictive relationship between the latent variables of individual and interpersonal functions and curiosity, with the interpersonal functions exercising a comparatively stronger influence. Prominent behavioral realizations of these functions are also discovered in a data-driven manner. We instantiate the proposed theoretical framework in a set of strategies and tactics that can be incorporated into learning technologies to indicate, evoke, and scaffold curiosity. This work is a step towards designing learning technologies that can recognize and evoke moment-by-moment curiosity during learning in social contexts and towards a more complete multimodal learning analytics. The underlying rationale is applicable more generally for developing computer support for other metacognitive and socio-emotional skills.
\end{abstract}


%
\IEEEpeerreviewmaketitle

\vspace{-0.2cm}
\section{Introduction}
\vspace{-0.05cm}
%
%
%
%
\IEEEPARstart{C}{uriosity} is defined as a strong desire to learn or know more about something or someone, and is an important metacognitive skill to prepare students for lifelong learning \hspace{-0.008cm}\cite{park2017tripartite}. Because curiosity is increasingly rare in students in classrooms, perhaps because increasingly rare in curricula \cite{engel2009curiosity,engel2011children}, large-scale attempts exist to help teachers bring back children's curiosity\footnote{https://tinyurl.com/TeachTeachersToEvokeCuriosity}. Traditional accounts of curiosity in psychology and neuroscience focus on how it can be evoked via underlying mechanisms such as novelty (features of a stimulus that have not yet been encountered), surprise (violation of expectations), conceptual conflict (existence of multiple incompatible pieces of information), uncertainty (the state of being uncertain), and anticipation of new knowledge \hspace{-0.008cm}\cite{jirout2012children,kidd2015psychology}. These knowledge-seeking experiences create a positive impact on student's beliefs about their competence in mastering scientific processes, in turn promoting greater breadth and depth of exploration \cite{wu2013modeling}.

Such theories have inspired the development of computer systems for facilitating task performance by enhancing an individual's curiosity (e.g., \cite{wu2013modeling,gordon2015can,law2016curiosity}), and simulating human-like curiosity in autonomous agents \cite{oudeyer2004intelligent}. Evoking curiosity here mainly focuses on directing an individual to a specific new knowledge component, followed by facilitating knowledge acquisition through exploration. Such a linear approach largely ignores how learning is influenced by social contexts. Here, a child's intrinsic motivation, exploratory behaviors, and subsequent learning outcomes may be informed not only by materials available to the child, but also the active work of other children, and the presence of facilitators \hspace{-0.008cm}\cite{parr2002environments,kashdan2004facilitating}. For instance, an expression of uncertainty or a hypothesis made by one child may cause peers to realize that they too are uncertain about the topic under discussion. This may initiate working together to overcome the cause of uncertainty, in turn having a positive impact on the individuals' curiosity \cite{jordan2014managing}.

Although substantial literature exists on the intrapersonal origins of curiosity, with rare exceptions (cf. \cite{engel2011children,wu2015modeling}) a lacuna exists concerning how peer-driven social factors contribute to moment by moment changes in curiosity. Peer influence is important because even in classrooms where teachers teach to the test, group work is encouraged, and such participation structures remain a core part of classrooms \cite{parr2002environments}. Thus, peers may be one of the few ways that curiosity is elicited. This makes it critical to understand curiosity beyond the individual level to an integrated knowledge-seeking phenomenon shaped by the individual, physical and social environment. Embodied Conversational Agents (ECAs) - particularly child-like ECAs called \textit{virtual peers} - have demonstrated a special capacity in supporting learning and collaborative skills for young children \cite{cassell2000shared}. Knowing how social factors influence curiosity allows the design of ECAs and other technologies to support curiosity-driven learning before children naturally support each other.

To address the goal of a framework that accounts for both individual and social influences on curiosity, our work takes a novel approach. We look at the impact of individuals on other individuals in group work by focusing on the observable behaviors that peers use - both language and nonverbal behavior - and using latent variable modeling to connect those observable behaviors to the students' inner curiosity states. Our work is unique in moving from a theoretical model to building arguments and evidence, relying on machine learning and human annotation, concerning verbal and nonverbal indicators that curiosity exists in a given student (and verbal and nonverbal behaviors that are most successful in eliciting curiosity, to building a statistical model that tests the theoretical assumptions.

More specifically, we propose an integrated socio-cognitive account of curiosity based on literature spanning psychology, learning sciences and group dynamics, and empirical analyses of a group informal learning environment. We make a bipartite distinction between functions that contribute to curiosity, and multimodal behaviors that fulfill these functions. Such functions are latent and therefore not directly knowable, and we must therefore hypothesize or presume them based on observable phenomena that index them. We therefore refer to them as putative functions. These functions comprise (i)``knowledge identification and acquisition” (helps humans realize that there is something they desire to know, and leads to the acquisition of the desired new knowledge), and (ii) ``knowledge intensification" (escalates the process of knowledge identification or acquisition by providing favorable environment) - at the individual and interpersonal level. We perform statistical validation of this framework to illuminate predictive relationships between multimodal behaviors, functions (latent variables because they cannot be directly observed), and ground truth curiosity (as judged by naive annotators). Longitudinal latent variable modeling is used to explicitly account for group structure and differentiate fine-grained behavioral dynamics.

The main contributions of this work are three-fold: First, it begins to fill the research gap of how social factors, especially interpersonal peer dynamics in group work, influence curiosity (section 6.B, 6.C). Second, the model is designed to lay a theoretical foundation to inform the design of learning technologies, a virtual peer in the current study, that can employ pedagogical strategies to evoke and maintain curiosity in social environments (section 6.C). Findings derived from the current analyses of human-human interaction can be informative in guiding the learning technology (human-agent interaction) design (section 7.A), and in decision making for multimodal analyses of behavior (section 7.B). Third, at the methodological level, our research (i) introduces novel approaches for collecting rich multimodal data in group settings (section 3.A), which is key to making fine-grained behavioral inferences, (ii) advances the use of crowdsourcing platforms for efficient ground truth annotation, which is important in human behavior analysis for educational research and beyond (section 5.A), (iii) provides a rigorous and reproducible semi-automatic behavior annotation approach, which combines complementary strengths of state-of-the-art machine learning methods and advantages of human judgment (section 5.B, 5.C and 5.D).

In what follows, Section 2 discusses related work. Section 3 discusses data collection (pre-studies and main study) across multiple learning contexts. Section 4 describes our combination of theory-driven and data-driven process for the development of the theoretical framework of curiosity. Section 5 describes the annotation of ground truth curiosity, verbal and nonverbal behaviors, and turn-taking metrics. Section 6 discusses statistical validation of the proposed theoretical framework of curiosity, with a discussion of the model fit to our corpus, identified causal interpersonal and intrapersonal behavior influence patterns that result in increase or maintenance of curiosity, and initial implementations of our theory-and-data driven approach in a virtual peer that is a co-player of a science game with children. In section 7, we describe the implications of this work for learning technology design and multimodal learning analytics. We end with limitations, future work and conclusion in sections 8, 9 and 10.

\vspace{-0.2cm}
\section{Related Work}
\vspace{-0.05cm}
For clarity, we divide related work into discussions about curiosity in the psychology, group dynamics and learning sciences literature and end with a brief overview of existing computational modeling approaches.

\vspace{-0.25cm}
\subsection{Curiosity in the Psychology Literature}
\vspace{-0.05cm}
Researchers in psychology describe curiosity as a psychological and behavioral state that ``responds to an inconsistency or a gap in knowledge" \cite{james1890}, and raises ``feelings of mystery, of strangeness, and of wonder" \cite{mcdougall1921}. Like hunger and thirst, curiosity is considered a critical internal drive for human beings that causes us to explore our environment, acquire knowledge and learn skills. It is generally described as an intrinsically motivated desire, passion or appetite for information, knowledge and learning \cite{kashdan2004facilitating,litman2005curiosity}. Several theoretical lenses explain the cause of curiosity. For instance, the incongruity theory argues that curiosity arises from the human tendency to make sense of the world on observing violated expectations \hspace{-0.008cm}\cite{piaget1969,hunt1963motivation}. Theories of conflict arousal consider curiosity as a drive that leads to the simultaneous occurrence of incompatible response tendencies. Main determinant factors for psychological conflict include both perceptual factors (e.g., novelty, surprise etc) and epistemic factors (e.g., incongruity, confusion etc) \cite{berlyne1960conflict}.

The information-gap perspective, another dominant view in the field, proposes that curiosity is raised when people attend to a gap in their knowledge, and the intensity depends on importance, salience and surprisingness of the desired information \hspace{-0.008cm}\cite{loewenstein1994psychology,golman2015curiosity,jirout2012children}. Other research in psychology has conceived of curiosity as desirable feelings associated with the anticipation of acquiring new knowledge. As described further below, our perspective adds to these causal factors verbal challenges, incompatible hypotheses, and negative evaluations of the target child's ideas
Previous literature also reveals several behavioral cues for curiosity. These include physical exploration such as orientation, locomotion and manipulation towards objects of interest, epistemic investigation such as question asking, experimentation, reasoning about observed phenomena, as well as expressions of surprise, excitement, wonder, confusion and attentiveness \hspace{-0.008cm}\cite{berlyne1960conflict,spektor2013science,baruch2016pre}.

Existing behavior-based measures of children's curiosity mainly rely on physical exploration and simple verbal behaviors such as question asking and commenting (e.g., \cite{zion2007curiosity,bonawitz2011double}). There is, however, a lack of rigorous operational measures of curiosity that incorporate the verbal and nonverbal behaviors displayed in real-time interaction.

\vspace{-0.25cm}
\subsection{Curiosity in the Group Dynamics Literature}
\vspace{-0.1cm}
Social accounts for curiosity remain largely unexplored in the psychology literature, and relevant research primarily focuses on parent-child or teacher-student interaction \hspace{-0.008cm}\cite{henderson1984parents,engel2011children}, instead of peer-peer interaction. Knowledge dissonance, social comparison and social information-seeking are three closely-knit factors related to curiosity and exploration which have been studied in prior work. First, compared to teacher-led learning where a teacher holds higher power/status in terms of knowledge possession, peer-peer learning is more likely to result in challenging different opinions or ideas from one another and active resolution of such knowledge incongruity or dissonance \cite{cartwright1953group}. This is one of the main sources of curiosity \cite{piaget1959language}. Controversy, instantiated as conflict or disagreement, has also been identified as one of the social causes of dissonance - the simultaneous existence of cognitions that in one way or another do not fit together. When individuals working in a small group experience dissonance, they might work towards reducing or eliminating dissonance via different means. One important means is attempt to seek additional social support for the held opinion by emphasizing its importance, lucrativeness, etc., in turn triggering curiosity \cite{cartwright1953group}.

Second, through social comparison, students are more eager to evaluate the correctness of their own opinion via group discussion and coming to know their peers' opinions, compared to knowing an expert's opinion \cite{radloff1961opinion}. When individuals face a question with no clear solution and they cannot reduce the uncertainty by consulting objective sources of information, they turn to views endorsed by others in the group and evaluate the accuracy of their beliefs by comparing themselves to others \cite{cartwright1953group}. Also, cognitive and affective changes are more likely when observing others who are perceived as friends or similar to the observer \cite{parr2002environments}. Also, students are more likely to actively seek information and solutions when their uncertainty is shared or at least considered as warranted, reasonable, or legitimate by their peers \cite{jordan2014managing}. Such joint hardship \cite{dornyei2003group} is likely to impact group member's behavior positively, due to the trigger it provides for engaging in cooperative/joint effort to overcome the obstacle by reasoning or physical exploration.


Third, social information seeking, or a general interest in gaining new social information (how others behave, act and feel) promotes acceptance (a non-evaluative feeling with unconditional positive regard towards another) \cite{dornyei2003group} and creates mutually shared cognition in the group \cite{van2006social}. It creates space for group members to learn from others' preferences and viewpoints \cite{engestrom1995polycontextuality}. Increased group member familiarity and knowledge awareness can increase willingness to work jointly and lead to consideration of more alternatives. Our research is meant to fill the research gap that exists concerning the role of individuals in social contexts, by incorporating theories from peer learning and group dynamics. Note that this perspective is quite separate from and complementary to more holistic accounts of group work, such as Cress' \cite{cress2008systemic} account of the co-evolution of collaborative knowledge, that describes the interplay between individual learning and group learning and its impact on group-level curiosity. Curiosity at the level of the group is surely at work in the contexts we analyzed as well, however, our focus on verbal and nonverbal behaviors produced by individuals and affecting other individuals' underlying psychological states precluded an analysis at the level of the group.

\subsection{Curiosity in the Learning Sciences Literature}
\vspace{-0.1cm}
Discussions about curiosity can be traced back to \cite{hatano1973intellectual}, who differentiated extensive curiosity (that widens a learner's interest) and particular curiosity (that helps a learner acquire detailed knowledge). \cite{ogata2000combining} tied these notions to the literature on knowledge awareness in collaborative learning settings. They posited that when a learner's activities are oriented towards the ``same knowledge" that the peer is looking at, discussing or changing, particular curiosity is excited. We find, on the contrary, that when a learner's activities are oriented towards ``different knowledge" than their peer, extensive curiosity can be evoked and collaboration possibilities are enhanced.

Curiosity has also been discussed under the umbrella of intrinsic motivation \hspace{-0.008cm}\cite{reiss2004multifaceted}. Intrinsically motivated learners derive pleasure from the task itself, while learners with extrinsic motivation rely on external rewards. \cite{keller1987strategies} considers curiosity as a motivational aspect in the design of learning technologies, and discusses surprising students as a central instructional tactic to lead them to explore new areas of the subject for constructing coherent explanations. Curiosity is also proposed as one of seven dimensions of the construct of ``learning power" \cite{shum2012learning}, which refers to a form of consciousness, or critical subjectivity leading to growth. Critically curious learners are comparatively less likely to accept what they are told uncritically, and more willing to reveal their questions and uncertainties.
It is important to note, however that, the focus of the learning sciences literature has been fundamentally cognitive, whereas we seek to understand the social scaffolding of curiosity along with its cognitive roots.

In the field of artificial intelligence in education, there is scarce research on how the dynamics of social interaction may influence student's intrinsic learning motivation. Some applications of curiosity can be seen in the development of pedagogical agents. For example, \cite{graesser1995collaborative} found that a curious peer (that keeps questioning) can problematize the interaction, and direct learner's attention to spot the contradiction in their knowledge structure, thereby inducing curiosity. \cite{wu2013innovative} also discovered that if a computer agent displays curiosity by pro-actively responding to novel conflicting stimuli, it can discover interesting learning concepts. Interaction with such a computer agent was shown to lead human learners to engage in greater task exploration of the learning environment, enhanced attention, and improved learning outcomes. The underlying model of interpersonal influence that is common to both these research strands is ``modeling". These systems implicitly assume human learners to spontaneously pick up on social cues of the ``curious'' pedagogical agent. We aim to develop a more nuanced understanding of curiosity, without directly equating behaviors to curiosity, or, relying solely on a single theoretical lens of looking at curiosity.

\vspace{-0.25cm}
\subsection{Computational Models of Curiosity}
\vspace{-0.1cm}
In general, curiosity has been computationally modeled \cite{wu2013curiosity} using an appraisal process where the incoming stimuli are first evaluated for their potential to provide an appropriate stimulation level. Subsequently, the stimulation level is mapped to a non-linear emotion curve called the Wundt curve \cite{wundt1874grundzuge} for deriving the curiosity level. The Wundt curve postulates too little stimulation to result in boredom, too much stimulation to result in anxiety, and only an optimal stimulation level to result in curiosity. We, however focus not just on perceptual, but also on the epistemic dimension of curiosity. Further, our computational model is meant to serve as an input into a human-pedagogical agent interaction in group work, unlike the goals of prior computational models of curiosity.

\vspace{-0.2cm}
\section{Data Corpus}
\vspace{-0.1cm}

\begin{figure*}[t]
\centering
\includegraphics[scale=0.57]{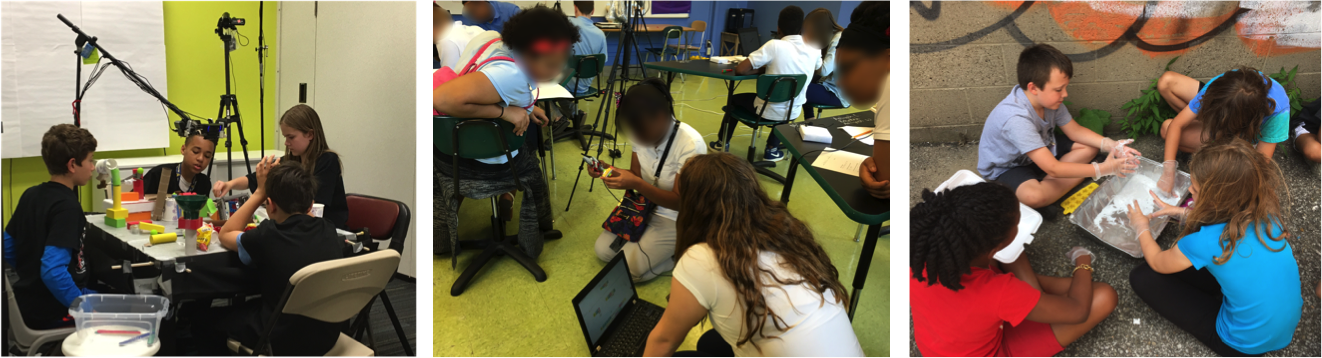}
\vspace{-0.2cm}
\caption{Empirical observation across learning contexts. Left: in-lab RGM building; Middle: in-school STEAM class; Right: science summer camp}
\vspace{-0.15cm}
\label{fig:empirical_studies}
\end{figure*}

\begin{figure*}[t]
\centering
\includegraphics[width=14.5cm]{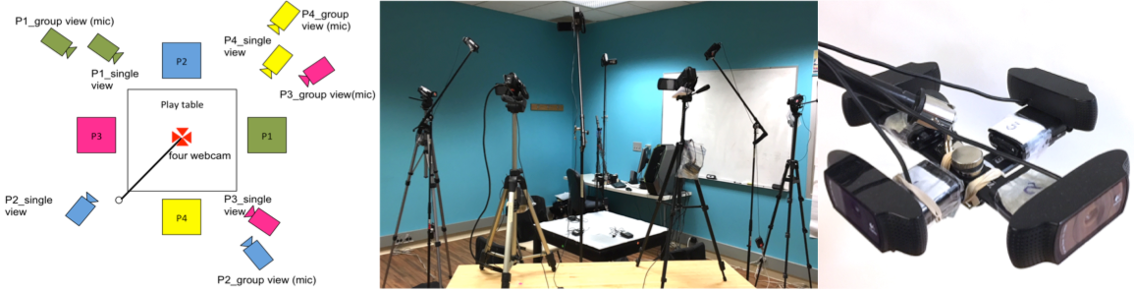}
\vspace{-0.2cm}
\caption{In-lab study data collection apparatus. Left: equipment arrangement; Middle: real arrangement; Right: fixture of four Webcam devices.}
\vspace{-0.15cm}
\label{fig:equipment setup}
\end{figure*}

\begin{figure*}[t]
\centering
\includegraphics[width=14cm]{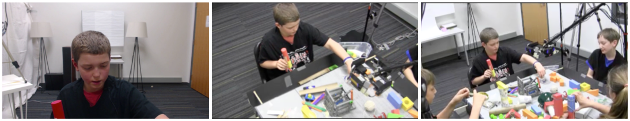}
\vspace{-0.2cm}
\caption{Illustration of video recording in the In-lab study. Left: frontal view; Middle: single view; Right: group view.}
\vspace{-0.45cm}
\label{fig:video recording}
\end{figure*}

To capture the simultaneous dynamic of fine-grained behaviors of multiple participants (Figure 1), a novel human behavior recording apparatus was developed. Recording equipment comprised four Logitech C920 Pro Webcam devices for the egocentric recording (frontal view), eight camcorder recorders for the exocentric recording (single and group view), and four lapel-microphones (Sennheiser ew100 G2 and Samson UHF Micro 32) to separate individual's speaking (Figure 2). We were able to get a frontal view of participants for automated detection of facial expression, head and gaze gesture, and single view and group view for analyzing other individual and social behaviors (Figure 3). Our goal in collecting such rich multimodal data was to be able to then annotate ground truth curiosity, detect verbal and nonverbal behaviors that fulfill putative functions of curiosity (see section 4.A, 4.B), and subsequently, statistically verify predictive relationships between them (see section 6).

\vspace{-0.25cm}
\subsection{Pre-study 1 (In-School)}
\vspace{-0.05cm}
Twenty participants in 6th grade (usually 11 to 12 years old) participated in our first pre-study. Participants were recruited by teachers in a local public school. There was no remuneration. The study took place in the 6th grade science class and STEAM class for two weeks. There were eight observation sessions in total, four sessions for science class and four sessions for STEAM class, with each session lasting about 45 minutes. The science class was focused on the topic of earth science learning, while the main activities of the STEAM class included robotics, programming, and crafting. Both classes were divided into small groups with 3-4 students. There were several constraints for data collection in a classroom, including large-scale observation (3-4 groups of students at the same time), noisy environments, and confined space and time for equipment set up. To accommodate the above constraints, we used a lightweight recording setup that included two camcorder recorders to capture students from the opposite sides of the table, and one lapel microphone in the middle of the table. Researchers also took field notes to describe student's interpersonal learning behaviors relating to curiosity.
\vspace{-0.27cm}

\subsection{Pre-study 2 (Science Summer Camp)}
\vspace{-0.05cm}
Thirty-one participants aged 9-14 participated in our second pre-study. Participants were recruited by teachers. The study took place in two science summer camp sessions hosted in a local child maker-space, with each summer camp lasting one week. We observed 30 hours in total. There was a wide variation of hands-on activities in the summer camp, spanning topics from physics, chemistry, biology, life science, robotics, and crafting. The class was divided into four to five groups. Compared to pre-study 1, the summer camp observation was longer and noisier, with more diverse (and less structured) activities. The class layout also changed frequently. We used one camcorder recorder and one lapel-microphone fixed on the table for each group of children. Similar to pre-study 1, researchers took field notes to describe peer-peer interaction associated with curiosity.
\vspace{-0.27cm}

\subsection{Main Study (In-Lab)}
\vspace{-0.05cm}
Our main study, which built upon the two pre-studies, was conducted in a semi-controlled adult-free lab setting. It comprised forty-four participants in 5th or 6th grade (16 male and 28 female, aged between 10-12, average age 11.2), who collaboratively built a Rube Goldberg machine (RGM) for about 35-40 minutes. The RGM task included building several chain reactions meant to be triggered automatically to trap a ball in the cage (without external human support). The machine was created using a variety of simple objects such as rubber bands, pipe cleaners, toy cars, clothespin, etc. We employed the RGM building activity because it supports learning of key science knowledge for students in 5th and 6th grade as defined by the Pennsylvania Department of Education \cite{SAS} such as force, motion and energy transfer, and enables collaborative hands-on learning and creative problem solving \cite{o2003application} without need for adult intervention.

Participants were recruited through flyers sent to local public schools, parent mailing-lists of the university, and advertisements on social media and in public spaces. All participants were remunerated with \$30 cash. 12 study sessions were videotaped, each one including a group of three to four students, and lasting for one and half hours. The procedure for each study session was as follows: (1) 10 minutes for an ice-breaking game; (2) 5 minutes free exploration of materials easily found at home; (3) 5 minutes of introducing participants to Rube Goldberg machines (RGM); (4) 30 minutes for collaboratively building the RGM; (5) a 5-minute opportunity to demonstrate the built RGM; (6) a 10-minute interview, and optionally (7) 5 minutes free exploration of a pre-made RGM; (8) 20 minutes for dispensing remuneration.
\vspace{-0.2cm}

\section{Theoretical Framework Development}
\vspace{-0.08cm}
We take a socio-cognitive view of curiosity whereby we acknowledge social influences but also try to isolate the individual mind as a cognitive unit of analysis by controlling for these external influences. This theoretical stance is partly borne out of our interest in studying individual curiosity (as opposed to a construct that is defined at the dyadic or group level like rapport). Adopting such a view also serves the practical goal of using behavioral cues of the other individuals to influence the curiosity of a given individual.

We define curiosity-driven learning in social contexts to be situations where particular forms of interaction among people trigger or facilitate behaviors associated with high curiosity. To leverage this notion for developing an integrated psychological and social framework of curiosity, a combination of theory-driven and data-driven approaches was used. From the theory-driven perspective, several iterations of the literature review were conducted with a gradual shift from individual- to interpersonal-level curiosity. Starting with research on curiosity at the individual level in psychology, we fanned out to critically investigate existing research on the social influence of curiosity within the group dynamics and the Learning Sciences literature. Informed by these theoretical lenses, we developed (i) a set of putative functions that contribute to curiosity, and (ii) multimodal behaviors that provide evidence for the potential presence of an individual's curiosity in the current time-interval because they fulfil these functions.

\vspace{-0.05cm}
From the data-driven perspective, we carried out empirical observation of small groups of 9-14 year old children engaged in hands-on learning activities across lab, science and STEAM class, and informal learning environments (section 3). Qualitative analyses were conducted in an exploratory manner first based on notes, videos and audio data collected during field observations \cite{ritchie2013qualitative}. The pre-studies helped in identifying a list of curiosity-related individual and interpersonal behaviors during small group science learning, based on our initial literature review. Building on these diverse pre-study contexts, we conducted follow-up qualitative/quantitative analyses on our in-lab data (but not constrained by the initial behavior list), to empirically validate and extend curiosity-related behaviors while minimizing confirmation bias.
\vspace{-0.05cm}

Overall, this mixed-methods approach involving the classroom, science summer camp, and lab as study contexts allowed us to (i) obtain familiarity to describe and explain the phenomenon of curiosity in different small group learning contexts, (ii) develop the integrated theoretical framework of curiosity with in-depth empirical evidence, and (iii) form relevant hypotheses based on the empirical exploration for follow-up quantitative validation of the theoretical framework.

We originally hoped to use both the pre-studies and main study data for quantitative analyses, but then discovered that adults in the classroom and summer program systematically obviated contexts in which curiosity might appear. The adults asked leading questions, rhetorical questions, and other kinds of classroom moves that left little room for exploration, creativity and curiosity. That in itself is an interesting finding, but one which we would like to address in subsequent work. While it is often said that classrooms represent more ecologically valid learning situations than do labs, we were interested in peer-peer interaction without adult intervention. Thus, in the current work, we looked at contexts in which curiosity might naturally occur in the absence of adults and examined the interpersonal triggers of that curiosity. For that reason, we also designed a lab context with open-ended activities where children were free to behave as they wished in the absence of adult supervision. This, we believe, was as close to ecologically valid as possible, given the presence of video cameras and adult (non-participating) observers.


\begin{table*} [t]
\centering
\caption{Corpus examples of behavior sequences. P1 is the child with high curiosity (see section 5.A for how we annotated curiosity)}\label{tab:behavior examples}
\scalebox{0.97}{
\begin{tabular}{|p{2.2cm}||p{7.55cm}||p{7.55cm}|} \hline
{\bf Behavior Cluster} & {\bf Empirical Observation (Example 1)} & \bf{Empirical Observation (Example 2)}\\ \hline

Cluster 1,2 & \pbox{7.55cm}{\scriptsize {\bf P1}: Hey let's..wait I have an idea \textit{[idea verbalization]} \\
{\bf P1}: Let's see what this is, but let me just, let me just.. \textit{[proposes joint action, co-occurs with physical demonstration, initiates joint inquiry]} \\
{\bf P2}: I have no idea how to do this, but it's making my
brain think \newline \textit{[positive attitude towards task]}} & \pbox{7.55cm}{\scriptsize {\bf P1}: So the chain has to be like this \textit{[idea verbalization with
iconic gesture]} \\
{\bf P1}: How would that be? \textit{[question asking followed by orienting towards stimulus]} \\
{\bf P1}: Well, I don't want it to break, so I want it to be about...no, let's say half an...half an inch \textit{[causal reasoning to justify actions being taken]}}\\ \hline \hline

Cluster 1,3 & \pbox{7.55cm}{\scriptsize {\bf P1}: Wait we need to raise it a bit higher \textit{[making suggestions]} \\
{\bf P1}: Maybe if we put it on..Umm..this thing maybe..this is high enough? \textit{[co-occurs with joint stimulus manipulation]} \\
{\bf P2}: Why? W-Why do we need to make it that high? \textit{[disagreement and asking for evidence]}} & \pbox{7.55cm}{\scriptsize {\bf P2}: And the funnel can drop it into one of um..those things \\
{\bf P1}: If the funnel can drop it… \\
{\bf P1}: Okay but then..even if it hits this, then we need what is this going to hit? \textit{[challenge]} \\
{\bf P1}: Here- let- just- make sure that it's going to hit it \textit{[followed by physical demonstration/verification]}} \\ \hline \hline

Cluster 2,3,4 & \pbox{7.55cm}{\scriptsize {\bf P1}: Roll off into here and go in there \textit{[hypothesis generation]} \\
{\bf P1}: Okay, so how are we going to do that? \textit{[question asking]} \\
{\bf P2}: It looks like something should hit the ball \textit{[making suggestion]}} & \pbox{7.55cm}{\scriptsize {\bf P2}: We could use this if we wanted \textit{[making suggestion]} \\
{\bf P1}: Let's figure this quickly...so we at least have this part done \textit{[preceded by expression of surprise and followed by trying to connect multiple objects to create a more complex object]}} \\ \hline

\end{tabular}}
\vspace{-0.2cm}
\end{table*}

\vspace{-0.3cm}
\subsection{Putative Functions Contributing to Curiosity}
\vspace{-0.1cm}
The iterative process described above led to the emergence of three function groups at the individual and interpersonal level. Because curiosity has traditionally been described as an inherently individual and stable disposition toward seeking novelty and approaching unfamiliar stimuli, we first outline individual aspects of curiosity for each function. We then flesh out interpersonal aspects of curiosity for every function. Each function can be realized in several different behavioral forms.

We call the first function group {\em Knowledge Identification}. As curiosity arises from a strong desire to obtain new knowledge that is missing or doesn't match with one's current beliefs, a critical precondition of this desire is to realize the existence of such knowledge. At an {\em individual} level, knowledge identification contributes to curiosity by increasing awareness of gaps in knowledge \cite{loewenstein1994psychology}, and highlighting relationships with related or existing knowledge to assimilate new information \cite{chi2014icap}. Further, exposure to novel and complex stimulus can raise uncertainty, subsequently resulting in conceptual conflict \hspace{-0.008cm}\cite{berlyne1960conflict,piaget1959language}. At an {\em interpersonal} level, knowledge identification contributes to curiosity by developing awareness of somebody else in the group having conflicting beliefs \cite{berlyne1960conflict} and awareness of the knowledge they possess \cite{ogata2000combining}, so that a shared conception of the problem can be developed \cite{van2006social}.

We call the second function group {\em Knowledge Acquisition}. Knowledge-seeking behaviors driven by curiosity not only contribute to the satisfaction of the initial desire for knowledge but also potentially lead to further identification of new knowledge. For example, question asking may help close one's knowledge gap by acquiring desired information from another peer. Depending on the response received, however, it may also lead to escalated uncertainty or conceptual conflict relating to the original question, thus reinforcing curiosity. At an {\em individual} level, knowledge acquisition involves finding sensible explanation and new inference for facts that do not agree with existing mental schemata \hspace{-0.008cm}\cite{schwartz2004inventing,chi2014icap}, and can be indexed by the generation of diverse problem-solving approaches \cite{schwartz2004inventing,sinha2020differential}. It also comprises comparison with existing knowledge or search for relevant knowledge through external resources to reduce simultaneous opposing beliefs stemming from the investigation \cite{cartwright1953group}. At an {\em interpersonal} level, knowledge acquisition comprises revelation of uncertainties in front of group members \cite{shum2012learning}, joint creation of new interpretations and ideas, engagement in argument to reduce dissonance among peers \cite{johnson2009energizing}, and critical acceptance of what is told \cite{shum2012learning}.

Finally, we call the third function group {\em Intensification of Knowledge Identification and Acquisition}. The intensity of curiosity or the desire for new knowledge is influenced by factors such as the confidence required to acquire it \cite{loewenstein1994psychology}, its incompatibility with existing knowledge and existence of a favorable environment \cite{cartwright1953group}. At an {\em individual} level, intensification of knowledge identification and acquisition can stem from factors such as anticipation of knowledge discovery \cite{dornyei2003group}, interest in the topic \cite{keller1987strategies}, willingness to try out tasks beyond ability without fear of failure \cite{sinha2020differential}, taking ownership of own learning and being inclined to see knowledge as a product of human inquiry \cite{shum2012learning}. These factors can subsequently result in a state of increased pleasurable arousal \cite{berlyne1960conflict}. At an {\em interpersonal} level, intensification of knowledge identification and acquisition is influenced by the willingness to get involved in group discussion and the tendency to be part of a cohesive unit in pursuit of instrumental objectives and/or for the satisfaction of a group member's affective needs \cite{cartwright1953group}.

Such willingness can span from the spectrum of merely continuing interacting to pro-actively reacting to the information others present \cite{van2006social}. Various interpersonal factors play out along this spectrum. Salient ones include interest in knowing more about a group member \cite{renner2006curiosity}, promotion of unconditional non-evaluative regard towards them \cite{dornyei2003group}, the tendency of spontaneous pickup of behavior initiated by a group member (where the initiator did not display any communicated intent of getting the others to imitate) \cite{cartwright1953group}, and awareness of one's own uncertainty being shared or considered legitimate by those peers \cite{jordan2014managing}. All these factors can result in a cooperative effort to overcome common blocking points for the group \cite{dornyei2003group}.

\vspace{-0.2cm}
\subsection{Behaviors Fulfilling Putative Functions of Curiosity}
\vspace{-0.1cm}
Behavioral episodes including language and associated multimodal communicative signals (e.g., facial expressions, gaze) serve as both communicative markers, i.e they provide evidence for the presence of curiosity of group members, and mind markers, i.e they shape group member's understanding and expectation of how to approach the task, along with conceptualization and construction of the associated knowledge \hspace{-0.008cm}\cite{heylen2006investigating}. They can (i) contain a single action or multiple co-occurring or contingent actions made by one or more individuals, (ii) be purposeful or non-purposeful because the underlying human strategy that governs the sequence of behaviors is unknown. Our review of prior research in psychology and learning sciences led us to link the behaviors with their putative functions in evoking curiosity, and organize these behaviors into four clusters. Table 1 illustrates examples of these behavior clusters from our empirical observations.

{\em Cluster 1} corresponds to behaviors that enable an individual to get exposed to and investigate physical situations, which may spur socio-cognitive processes beneficial to curiosity-driven learning \hspace{-0.008cm}\cite{berlyne1960conflict,chi2014icap}. Examples include orientation (using eye gaze, head, torso) and interacting with stimuli (e.g., manipulation of objects). When looking at video segments tagged with high curiosity in our empirical data, these behaviors occur in contexts where children look at different aspects of the stimulus (e.g., the function of novel objects, physical properties of mineral samples in the science class, transition phase of dry ice samples in the summer camp, etc) by orienting towards it using their gaze and torso, smelling or scratching it, rotating and trying to fit more than one object together, etc. {\em Cluster 2} corresponds to behaviors that enable an individual to actively make meaning out of their observations \hspace{-0.008cm}\cite{berlyne1960conflict,luce2015science,chi2014icap}. Examples include idea verbalization, justification and generating hypotheses. {\em Cluster 3} corresponds to behaviors that involve joint investigation with other peers \hspace{-0.008cm}\cite{berlyne1960conflict,luce2015science,chi2014icap}. Examples include arguing, evaluating the problem-solving approach of a partner (positive or negative), expressing disagreement, making suggestions, sharing findings, question asking. In video segments tagged with high curiosity, these behaviors occur in contexts where children listen to other's suggestions, express disagreement or challenge their perspective by pointing out loopholes, and engage in a physical demonstration for clarification. Finally, {\em Cluster 4} corresponds to behaviors that reveal an individual's affective states \hspace{-0.008cm}\cite{mcdaniel2007facial,kashdan2004facilitating} including surprise, enjoyment, confusion, uncertainty, flow and sentiment towards task.

We hypothesize that behaviors across these clusters will map onto one or more putative functions of curiosity, because there can be many different functions or reasons why a communicative behavior occurs. For example, in knowledge-based conflict in group work, attending to differing responses of others compared to one's own may raise simultaneous opposing beliefs {\em (knowledge identification)}. This awareness might in turn activate cognitive processes, wherein an individual may seek social support for one's original belief by emphasizing its importance and validating one’s idea by providing justification, or, engaging in a process of back and forth reasoning to come to a common viewpoint {\em (knowledge acquisition)}. Further, this awareness may as well impact socio-emotional processes, where an individual may perceive a conflict differently and their emotions felt and expressed might vary depending on relation with and perception of the source of conflict, e.g., is it a more competent/less competent, more cooperative/less cooperative peer that raises conflict, and therefore take the next action of resolving that conflict differently {\em (intensification of knowledge identification and acquisition)}. We intend to discover prominent mappings between functions described in section 4.A and behaviors described in section 4.B more formally in a data-driven way in section 6.
\vspace{-0.2cm}

\section{Quantitative Analyses of In-Lab Study}
\vspace{-0.05cm}
We now describe fine-grained quantitative analyses from a convenience sample of the first 30 minutes (out of 35-40 minutes given each group), of the RGM task (lab study) for half of the sample; that is, 22 children across 6 groups. Table 2 provides a summary of all coding metrics used in this article. Our goal is to empirically verify the theoretical framework of curiosity proposed in section 4.

\begin{table*} [t]
\vspace{-0.1cm}
\caption{Summary of coding methods (detailed coding scheme at http://www.tinyurl.com/codingschemecuriosity)}\label{tab:coding}
\centering
\scalebox{0.69}{
\begin{tabular}{|p{3.0cm}|p{17cm}|p{4.8cm}|} \hline
{\bf Construct}
& {\bf Definition used to code/infer the construct} & {\bf Coding method}\\ \hline
\pbox{2.5cm}{Ground Truth\\ Curiosity} & A strong desire to learn or know more about something or someone. & Four MTurk raters annotated each 10-sec thin slice; average ICC = 0.46; used inverse-based bias correction to pick the final rating. \\ \hline \hline

\multicolumn{3}{|l|}{\bf Verbal Behavior}\\\hline

1. Uncertainty & Lack of certainty about ones choices or beliefs, and is verbally expressed by language that creates an impression that something important has been said, but what is communicated is vague, misleading, evasive or ambiguous. \newline e.g - {\em ``well maybe we should use rubberbands on the foam pieces", ``wait do we need this thing to funnel it through?"} \newline & \multirow{5}{*}{\pbox{4.8cm}{Used a semi-automated annotation approach: after automatic labeling of these verbal behaviors, two trained raters (Krippendorff's alpha $>$ 0.6) independently corrected machine annotated labels; average percentage of machine annotation that remained the same after human correction was 85.9 ($SD=$ 12.71). }} \\
\hhline{--~}2. Argument &A coherent series of reasons, statements, or facts intended to support or establish a point of view. \newline e.g -{\em ``no we got to first find out the chain reactions that it can do", ``wait, but anything that goes through is gonna be stuck at the bottom"} \newline & \\
\hhline{--~}3. Justification & The action of showing something to be right or reasonable by making it clear. \newline e.g -{\em ``oh we need more weight to like push it down", ``wait with the momentum of going downhill it will go straight into the trap"} \newline &\\
\hhline{--~}4. Suggestion&An idea or plan put forward for consideration. \newline e.g - {\em ``you could kick a ball to kick something", ``you are adding more weight there which would make it fall down"} \newline & \\
\hhline{--~} 5. Agreement & Harmony or accordance in opinion or feeling; a position or result of agreeing. \newline e.g - {\em ``But we need to have like power, and weight too" {\em (Quote)} --- ``Yeah we need more weight on this side" {\em (Response)}, ``And we put the ball in here..I hope it still works, and it goes..so it starts like that, and then we hit it" {\em (Quote)} --- ``Ok that works" {\em (Response)}} & \\\hline \hline

6. Question Asking\newline (On-Task/Social) & Asking any kind of questions related to the task or non-task relevant aspects of the social interaction. \newline e.g - {\em ``so what's gonnna..what will happen like after the balls gets into the cup?", ``why do we need to make it that high?", ``do you want to build something like a chain reaction or something like that?", ``do you two go to the same school?", ``who else watched the finale of gravity falls?"}& \multirow{6}{*}{\pbox{4.8cm}{Used manual annotation procedure due to unavailability of existing training corpus (Krippendorff's alpha $>$ 0.76 between two raters).}} \\

\hhline{--~}7. Idea Verbalization & Explicitly saying out an idea, which can be just triggered by an individual's own actions or something that builds off of other peer's actions. \newline e.g - {\em ``yeah that ball isn't heavy enough", ``so it's like tilted a bit up so it catches it instead of tilted down"} & \\
\hhline{--~}8. Sharing Findings& An explicit verbalization of communicating results, findings and discoveries to group members during any stage of a scientific inquiry process. \newline e.g - {\em ``look how I'm gonna see I'm gonna trap it", ``look I made my pillar perfect"}& \\
\hhline{--~}9. Hypothesis\newline Generation &Expressing one or more different possibilities or theories to explain a phenomenon by giving relation between two or more variables. \newline e.g - {\em ``we could use scissors to cut off the baby's head which would cause enough friction", ``okay we need to make it straight so that the force of hitting it makes it big"} & \\
\hhline{--~}10. Task Sentiment\newline (Positive/Negative)&A view of or attitude (emotional valence) toward a situation or event; an overall opinion towards a subject matter. We were interested in looking at positive or negative attitude towards the task that students were working on. \newline e.g - {\em ``oh it's the coolest cage I've ever seen, I'd want to be trapped in this cage", ``ok so I'm gonna try to find out a way for the end to make this one go and fall",``I'm getting very mad at this cage",``but I don't know how to make it better"} & \\
\hhline{--~}11. Evaluation \newline (Positive/Negative)&Characterization of how a person assesses a previous speaker's action and problem-solving approach. It can be positive or negative. \newline e.g - {\em ``oh that's a pretty good idea - that was a good idea",``let's make this thing elevated and make it go down",``oh wait this doesn't- you're not pushing anything over here", ``no it can't go like that otherwise it will be stuck"} & \\\hline\hline

\multicolumn{3}{|l|}{\bf{Nonverbal Behavior (AU - facial action unit)}} \\\hline

1. Joy-related & AU 6 (raised lower eyelid) {\em and} AU 12 (lip corner puller). & \multirow{5}{*}{\pbox{4.8cm}{Used an open-source software OpenFace for automatic facial landmark detection, and a rule-based approach post-hoc to infer affective states}} \\
\hhline{--~}2. Delight-related & AU 7 (lid tightener) {\em and} AU 12 (lip corner puller) {\em and} AU 25 (lips part)
{\em and} AU 26 (jaw drop) {\em and not} AU 45 (blink). & \\
\hhline{--~}3. Surprise-related & AU 1 (inner brow raise) {\em and} AU 2 (outer brow raise) {\em and}
AU 5b (upper lid raise) {\em and} AU 26 (jaw drop). & \\
\hhline{--~}4. Confusion-related & AU 4 (brow lower) {\em and} AU 7 (lid tightener) {\em and not} AU 12 (lip corner puller). & \\
\hhline{--~}5. Flow-related & AU 23 (lip tightener) {\em and} AU 5 (upper lid raise) {\em and} AU 7 (lid tightener)
{\em and not} AU 15 (lip corner depressor) {\em and not} AU 45 (blink) {\em and not} AU 2 (outer brow raise). & \\ \hline \hline
\hhline{--~}6. Head Nod & Variance of head pitch. & \multirow{3}{*}{\pbox{4.8cm}{Used OpenFace to extract head orientation, and computed variance post-hoc}} \\
\hhline{--~}7. Head Turn & Variance of head yaw. & \\
\hhline{--~}8. Lateral Head \newline Inclination & Variance of head roll.& \\\hline\hline

\multicolumn{3}{|l|}{\bf{Turn Taking}} \\\hline
1. Indegree & A weighted product of number of group members whose turn was responded to ({\em activity}) and total time that other people spent on their turn before handing over the floor ({\em silence}). & \multirow{2}{*}{\pbox{4.8cm}{Used two novel metrics constructed using an application of social network analysis for weighted data.} } \\
\hhline{--~}2. Outdegree & A weighted product of number of group members to whom floor was given to ({\em participation equality}), and the amount of time spent when holding floor before allowing a response ({\em talkativeness}). & \\\hline

\end{tabular}}
\vspace{-0.23cm}
\end{table*}
\vspace{-0.27cm}

\subsection{Assessment of Ground Truth Curiosity}
\vspace{-0.1cm}
Person perception research has demonstrated that judgments of others based on brief exposure to their behaviors give an accurate assessment of interpersonal dynamics \cite{ambady1992thin}. We used Amazon's Mechanical Turk (MTurk) platform to obtain ground truth for curiosity via such a thin-slice approach, using the definition ``curiosity is a strong desire to learn or know more about something or someone", and a rating scale comprising 0 (not curious), 1 (curious) and 2 (extremely curious). Crowdsourcing platforms provide benefits of a diverse sample of raters who can be accessed quickly, easily and for relatively little cost \cite{follmer2017role}. Our previous research has successfully deployed thin-slice coding for other social phenomena such as interpersonal rapport in peer tutoring using MTurk \hspace{-0.008cm}\cite{sinha2015we,zhao2016socially}.

Here, four na{\"i}ve raters annotated every 10-second slice of videos of the interaction for each child presented in randomized order. We post-processed ratings by removing raters who used less than 1.5 standard deviation time compared to the mean time taken for all rating units (HITs). We then computed a single measure of Intraclass correlation coefficient (ICC) for each possible subset of raters for a particular HIT, and then picked ratings from the rater subset that had the best reliability for further processing. Finally, inverse-based bias correction \cite{kruger2014axiomatic} was used to account for label overuse and underuse, and to pick one single rating of curiosity for each 10-second thin-slice. The average ICC of 0.46 aligns with reliability of curiosity in prior work \hspace{-0.008cm}\cite{nojavanasghari2016emoreact,craig2008emote}.
\vspace{-0.2cm}

\subsection {Assessment of Verbal Behaviors}
\vspace{-0.1cm}
We adopted a mix of semi-automatic and manual annotation procedures to code 11 verbal behaviors, in line with the curiosity-related behavioral set described in section 4.B. These verbal behaviors span propositional \cite{cassell2003negotiated} and interpersonal \cite{zhao2016automatic} functions of contributions to a conversation. Propositional functions are those that are fulfilled by contributing informational content to the dialog (e.g., idea verbalization, justification), and interpersonal functions are those that are fulfilled by managing the relationship between the interlocutors (e.g., social question asking, positive evaluation). Five verbal behaviors were coded using a semi-automatic approach - {\em uncertainty, argument, justification, suggestion} at the clause level, and {\em agreement} at the turn level. A clause contains a subject (a noun or pronoun) and a predicate (conjugated verb – that says something about what the subject is or does). During a full turn, a speaker holds the floor and expresses one or more interpretable clauses.

First, a particular variant of neural language models called paragraph vector or doc2vec \cite{le2014distributed} was used to learn distributed representations for a clause/turn. This means that for every clause/turn in our data corpus, we transformed the sequence of words in it to a tuple (or vector) of continuous-valued features that characterize the semantic meaning of those words. Such feature representation implies that sentences in a test set that functionally similar to sentences in a training set can still achieve good predictions. The motivation for this approach stems from: (i) lack of available corpora of verbal behaviors that are large enough, and collected in similar settings as ours (groups of children engaged in scientific inquiry), and hence (ii) limited applicability of traditional n-gram based models to cross-domain settings, which would result in a high-dimensional representation with poor semantic generalization, (iii) limitations of other neural language models such as word2vec that do not explicitly represent word order and surrounding context (in contrast, doc2vec models contain an additional paragraph token that acts as a memory and remembers what is missing from the current context, thus not ignoring word order that is important for the semantic representation), and (iv) our desire to reduce manual annotation time when each child's behaviors must be annotated.
\vspace{-0.15cm}

Based on the recommended procedure in \cite{le2014distributed}, we used concatenated representations of two fixed size vectors of size 100 that we learned for each sentence as input to a machine learning classifier (L2 regularized logistic regression) - one learned by the standard paragraph vector with distributed memory model, and one learned by the paragraph vector with a distributed bag of words model. Empirically too, we found this concatenated vector representation to perform better on cross-validation performance on the training data, compared to using any of the two vector representations alone. Training data for the five verbal behaviors annotated using this process is shown in the right column of Table 3, along with standard performance metrics such as weighted F1 score (to account for class imbalance) and Area under ROC curve (AUC). Test data comprised the in-lab study corpus (section 4.C).
\vspace{-0.15cm}

\begin{table*}
\centering
\caption{Results from semi-automatic verbal behavior annotation. Right column describes external corpus used for training machine learning classifiers \& their performance. Left column depicts inter-rater reliability for human judgment used in verifying robustness of machine annotated labels.}\label{tab:1}
\scalebox{0.73}{
\begin{tabular}{|p{8cm}|p{16cm}|} \hline
{\bf Verbal Behavior [Krippendorff's $\alpha$ for human judgment]}
& {\bf Training Data for Semi-Automated Classification \{Weighted F1, AUC\} (10-fold cross validation)]} \\ \hline
1. Uncertainty [0.78]
& Wikipedia corpus manually annotated for 3122 uncertain 7629 certain instances \cite{farkas2010conll} \{0.695, 0.717\}\\ \hline
2. Argument [0.792]
& Internet Argument Corpus manually annotated for 3079 argument and 2228 non argument instances \cite{swanson2015argument}. Argument quality score split at 70\% to binarize class label
\{0.658, 0.706\}\\ \hline
3. Justification
[process (0.936), causal (0.905), model (0.821), example (0.731), definition (0.78), property (0.847)]
& AI2 Elementary Science Questions corpus manually annotated for 6 kinds of justification - process, causal, model, example, definition, property \cite{jansens}. Reported performance is the average performance of 6 binary machine learning classifiers
\{0.766, 0.696\}\\ \hline
4. Suggestion [0.608]
& Product reviews \cite{negi2016suggestion} and Twitter \cite{dong2013automated} corpuses manually annotated for 1000 explicit suggestion and 13000 explicit non-suggestion instances
\{0.938, 0.865\}\\ \hline
5. Agreement [0.935] \newline
& LiveJournal forum and Wikipedia discussion corpuses manually annotated for 2754 agreement and 8905 disagreement instances based on quote and response pairs \cite{andreas2012annotating} \{0.717, 0.696\}\\ \hline


\end{tabular}}
\vspace{-0.2cm}
\end{table*}

The robustness of machine annotated labels on the test data was ensured using human annotators. Two raters first coded presence or absence of verbal behaviors on a random sample of 100 clauses/turns following a coding manual given to them for training, and computed inter-rater reliability using Krippendorff's alpha. Once raters reached a reliability of $>$ 0.7 after one or more rounds of resolving disagreements, they independently rated a different set of 50 clauses/turns independently, and we computed the final reliability on these (left column of Table 3, and $>$ 0.6 for all behaviors). Subsequently, the raters independently de-noised or corrected machine annotated labels for the full corpus, and we use these final labels for empirical validation of the theoretical framework (as described in section 6). Compared with this human ground truth, the average ratio of false positives to false negatives across all annotation categories in the machine prediction was 14.18 ($SD=$ 12.31), suggesting over-identification of the presence of verbal behaviors.

We found that the most common false positives were cases where a clause or turn comprised one word (e.g., okay), backchannels (e.g., - hmmm..) and very short phrases lacking enough context to make a correct prediction. The average percentage of machine annotated labels that did not change even after the human de-noising step was 85.9 ($SD=$ 12.71). This meant that majority of the labels were correctly predicted in the first place. This was also reflected in a good cross validation training performance of the models (right column of Table 3). Six other verbal behaviors ({\em question asking (on-task, social) ($\alpha=$ 1), idea verbalization ($\alpha=$ 0.761), sharing findings ($\alpha=$ 1), hypothesis generation ($\alpha=$ 0.79), attitude towards task (positive, negative) ($\alpha=$ 0.835), evaluation sentiment (positive, negative) ($\alpha=$ 0.784)}) were coded using manually due to unavailability of existing training corpus.

\vspace{-0.2cm}
\subsection{Assessment of Nonverbal Behaviors}
\vspace{-0.1cm}
Our current work focused initially on the nonverbal behaviors of the face, operationalized as facial landmarks, and based on a rich body of literature describing movements of the face in learning \cite{cahour2013characteristics}, and particularly the facial action units that index certain emotions that often co-occur with curiosity \cite{nojavanasghari2016emoreact}. This work has discovered consistent associations (correlations as well as predictions) between particular facial configurations and human emotional or mental states \hspace{-0.008cm}\cite{mcdaniel2007facial,grafsgaard2011modeling,nojavanasghari2016emoreact}.
We used automated visual analysis to construct five feature groups corresponding to emotional expressions that provide evidence for the presence of the affective states of {\em joy, delight, surprise, confusion} and {\em flow} (a state of engagement with a task such that concentration is intense). A simple rule-based approach was followed (see Table 2) to combine emotion-related facial landmarks, which were previously extracted on a frame by frame basis using a state-of-the-art open-source software OpenFace \cite{baltruvsaitis2016openface}. We then selected the most dominant (frequently occurring) emotional expression for every 10-second slice of the interaction for each group member, among all the frames in that time interval.

Automated visual analysis was also used to capture variability in head angles for each child in the group, which correspond to {\em head nods (i.e. pitch), head turns (i.e. yaw)}, and {\em lateral head inclinations (i.e. roll)}. The motivation for using head movement in our curiosity framework is inspired by our prior work in nonverbal behavior and learning technologies \cite{ryokai2002literacy} and in multimodal analytics \hspace{-0.008cm}\cite{gatica2005detecting,schuller2009being} which have emphasized the contribution of nonverbal cues in inferring behavioral constructs such as interest and involvement that are closely related to the construct of curiosity, and which have demonstrated the positive impact of nonverbal behavior by virtual peers in children's learning. By using OpenFace\cite{baltruvsaitis2016openface}, we first performed frame by frame extraction of head orientation and then calculated the variance post-hoc to capture intensity in head motions for every 10 second of the interaction for each group member. Because head pose estimation takes as input facial landmark detection, we only considered those frames that had a face tracked and facial landmarks detected with confidence greater than 80\%.
\vspace{-0.35cm}

\subsection{Assessment of Turn Taking Dynamics}
\vspace{-0.15cm}
While the annotated verbal behaviors fulfill propositional and interpersonal conversational goals in the social interaction, the interactional function of contributions to a conversation is captured by turn-taking behaviors. Interactional discourse functions are ``responsible for creating and maintaining an open channel of communication between the participants" \cite{cassell2003negotiated}. The motivation for capturing turn taking in the current research stems Further from prior literature that has used measures such as participation equality and turn taking freedom as indicators of involvement in small-group interaction \cite{lai2013detecting}.

In the current work, we designed two novel metrics using a simple application of social network analysis for weighted data. By representing speakers as nodes and time between adjacent speaker turns as edges, we computed two features for each group member. These features calculated for every 10 seconds, comprised (i)
{\em TurnTakingIndegree}, a weighted product of the number of group members whose turn was responded to (activity) and the total time that other people spent on their turn before handing over the floor (silence), and was quantified as {\em activity}$^{1-\alpha}$ $*$ {\em silence}$^{\alpha}$. Because high involvement is likely to be indexed by higher activity and lower silence, $\alpha$ was set to -0.5, (ii)
{\em TurnTakingOutdegree}, a weighted product of the number of group members to whom floor was given (participation equality), and the amount of time spent when holding floor before allowing a response (talkativeness), and was quantified by {\em participation equality}$^{1-\alpha}$ $*$ {\em talkativeness}$^{\alpha}$. Because higher participation equality and talkativeness are favorable, $\alpha$ was set to +0.5. These two metrics were used in our empirical validation.
\vspace{-0.35cm}

\section{Empirical Validation of the Proposed Theoretical Framework of Curiosity}
\vspace{-0.05cm}
We used a ``multiple-group" version of continuous time structural equation model (CTSEM) \cite{drivercontinuous} to evaluate the proposed theoretical framework of curiosity, and statistically verify the predictive relationships between ground truth curiosity (formalized as a manifest variable), putative functions described in our theoretical framework (formalized as latent variables) and multimodal behaviors (formalized as time-dependent predictors). For clarity, we first present a rationale for the choice of our modeling framework, and then a formal description of the technical background underlying these latent variable analyses. Finally, we describe a concrete application of this modeling framework to our in-lab study corpus.
\vspace{-0.25cm}

\subsection {Modeling Framework (CTSEM)}

\subsubsection{Rationale}
Conventional Structural Equation Models (SEMs) assume independence of observations, and thus cannot be applied directly to analyze autocorrelated time series data from multimodal behavior analyses. This suggests consideration of a Dynamic Bayesian Network-like model to explicitly model temporal dependencies between the latent random variables across time-steps. Applications of such models in the social and behavioral sciences are usually limited to discrete time models, with the assumption that time progresses in discrete steps, and intervals between measurement occasions are equal. In many cases, these assumptions are not met, resulting in biased parameter estimates and a misunderstanding of the strength and time course of effects. Continuous time SEM models overcome these limitations by using multivariate stochastic differential equations to estimate an underlying continuous process and recover underlying latent or hidden causes linking the entire sequence.


\subsubsection{Technical Description}
To clarify how CTSEM improves over SEM models in the current context, we give a short technical description here. Formally, a multivariate stochastic differential equation for a latent process of interest in CTSEM can be written as d$\eta_i(t) = A\eta_i(t) + Bz_i + M\chi_i(t) + GdW_i(t) + \xi_i$ {\em (Structural part of the SEM model)}, where $A$ is the drift matrix that models auto effects the latent variable has on itself on the diagonals, and cross effects to other latent processes on the off-diagonals, in turn characterizing temporal relationships between the processes. $\xi_i$ determines the long-term level of the latent process. Matrix $B$ and $M$ represent the effect of time-independent and time-dependent variables on the latent process. Time-independent predictors would typically be variables that differ between subjects, but are constant within subjects for the time range in question (e.g., a trait curiosity questionnaire). Time-dependent predictors vary over time and are independent of fluctuations of the latent processes in the system. They can be treated as a simple impulse form where the predictors are treated as impacting these processes only at the instant of an observation.

The matrix $G$ represents the effect of noise or the stochastic error term $dW_i(t)$ on the change in the latent process. $Q$ = $G * G^T$ represents the variance-covariance matrix of diffusion process in continuous time. The essence of diffusion processes is to capture very slow patterns of change in the latent variable. Further, this latent process can be used to predict manifest variables of interest using the equation $y_i(t)=\Lambda \eta_i(t) + \zeta_i(t)$ {\em (Measurement part of the SEM model)}, where $\Lambda$ is a matrix of factor loadings between the latent and manifest variables and $\zeta_i$ is the residual (error) vector.

A Kalman filter can be used to fit CTSEM to the data and obtain standardized estimates for the influence of behaviors on latent functions, and in turn these latent functions on curiosity. It uses a series of measurements observed over time (containing statistical noise and other inaccuracies) to produce estimates of unknown variables that tend to be more precise than those based on a single measurement alone. It is a state space model described by a (i) state equation that describes how the latent states change over time and is analogous to {\em structural part of the SEM model}, and (ii) output equation that describes how the latent states relate to the observed states at a single point in time (how the observed output is produced by the latent states), and is analogous to {\em measurement part of the SEM model.}
In the presence of multiple groups in a dataset (e.g., we have 6 groups in our corpus), a ``multiple group" version of CTSEM should be used. It allows investigation of group level differences and helps understand variability in model parameters across different groups.

\begin{figure*}[t]
\centering
\scalebox{0.88}{
\includegraphics[width=\textwidth]{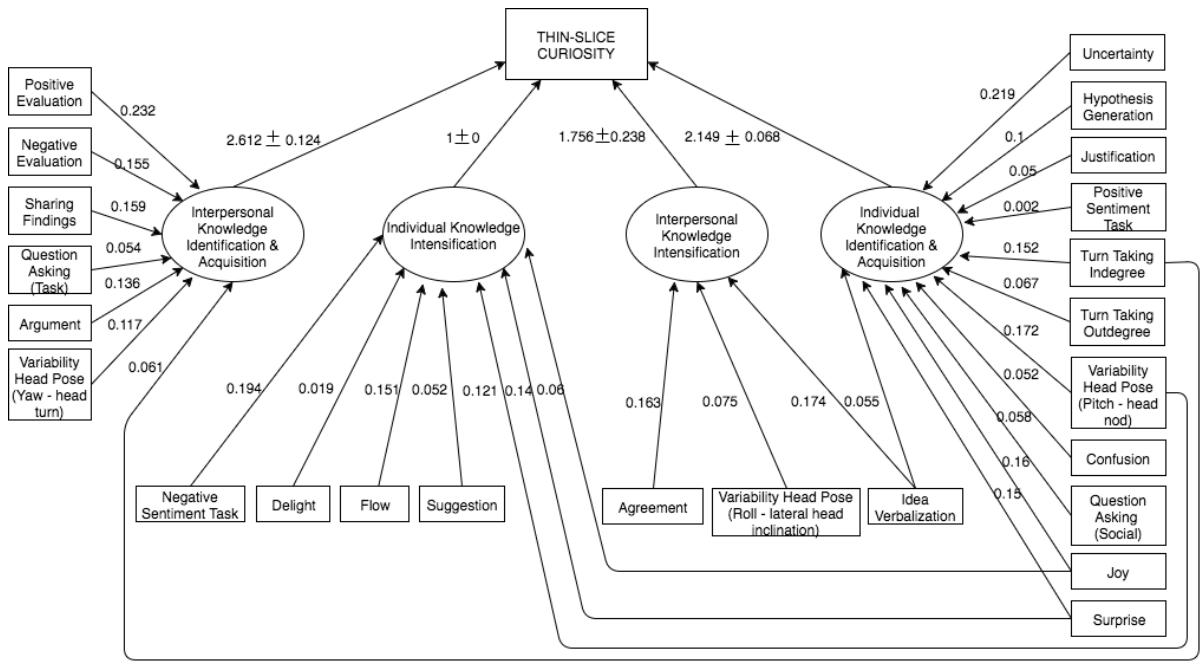}}
\vspace{-0.2cm}
\caption{Empirical validation of the theoretical framework of curiosity depicting fit of the Continuous time structural equation model (CTSEM) to the In-Lab study corpus. Rectangles represent observed constructs, while ovals represent latent constructs. Direction and degree of predictive influences are represented by edges between multimodal behaviors (time dependent predictors), putative functions of curiosity (latent variables) and thin-slice curiosity (manifest variable). Degree of predictive influence between latent and manifest variable is averaged across 6 study groups.}
\vspace{-0.2cm}
\label{fig:SEM}
\end{figure*}

\subsubsection{Application to the In-Lab Study Corpus}
Because knowledge identification and acquisition are closely intertwined with knowledge-seeking behaviors and it is hard to distinguish between these underlying mechanisms based on observable or inferred multimodal behaviors, we formalized them under the same latent variable. The final set of latent functions that we statistically verified therefore included: (i) {\em individual} knowledge identification and acquisition, (ii) {\em interpersonal} knowledge identification and acquisition, (iii) {\em individual} intensification of knowledge identification and acquisition, (iv) {\em interpersonal} intensification of knowledge identification and acquisition. We ran two versions of CTSEM.

In the first version, we specified a model where only factor loadings between the manifest variable and latent variables in {\em measurement part of the model} were estimated for each group distinctly (we report the average and standard deviation across the 6 groups in Figure 4), but all other model parameters including those belonging to {\em structural part of the model} were constrained to equality across all groups (Model$_{constrained}$) and then estimated freely. This means that matrices $A$, $B$, $M$, $G$ and $\Lambda$ were freely estimated. Because the form of a behavior does not uniquely determine its function, nor vice-versa, we did not pre-specify the exact pattern of relationships between behaviors and functions to look for/estimate. In the second version of the model, all parameters for all groups were estimated distinctly (Model$_{free}$).

Technically, the rich representational capacity of ``multiple-group" CTSEM allows running these two separate models. However, analytically, the decision to separately run these two models was based on the intuition that while the relationships between the appearance of behaviors and their contribution to the latent functions of curiosity would remain the same across groups, the relative contribution of interpersonal or individual tendencies for knowledge identification, acquisition and intensification would vary based on learning dispositions of people towards seeking the unknown. This intuition stemmed from prior literature that has looked into measuring learning dispositions \cite{shum2012learning}, an important dimension of which is the ability of learners to balance between being sociable and being private in their learning, i.e not being completely independent or dependent, but rather working interdependently. We hypothesized that this dimension will impact curiosity differently, especially because our data comprised a group learning context, and therefore expected Model$_{constrained}$ to fit the data better than Model$_{free}$.
\vspace{-0.2cm}

\subsection{Model Results and Discussion}
\vspace{-0.05cm}
An empirical validation confirmed our hypothesis of Model$_{constrained}$ fitting to the data better than Model$_{free}$. The Akaike Information Criterion (AIC) for Model$_{constrained}$ (933.48) was $\sim$3x lower than Model$_{free}$ (2278.689). We now illustrate results of the CTSEM (Model$_{constrained}$) in Figure 4, depicting links with top-ranked standardized estimates between behaviors and latent variables. In few cases, we also added links with the second-highest standardized estimate if they clarified our interpretation of the latent function. Because a latent variable is unmeasured, its units of measurement must be fixed by a researcher for the model to be identified (for there to exist a unique solution for all of the model's parameters). Therefore, one factor loading was conventionally fixed to 1.

Overall, these results confirm the correctness of our framework along three main aspects: (i) The grouping of behaviors under each latent function and their contribution to individual and interpersonal aspects of knowledge identification, acquisition and intensification aligns with prior literature on the intrapersonal origins of curiosity, but also teases apart the underlying interpersonal mechanisms, (ii) There exists strong and positive predictive relationships between these latent variables and thin-slice curiosity, (iii) Knowledge identification and acquisition have a stronger influence to curiosity than knowledge intensification, and interpersonal-level functions have stronger influence compared to individual-level functions. We now discuss latent functions and associated behaviors, ordered by the degree of positive influence on curiosity.

First, {\em Interpersonal Knowledge Identification and Acquisition} shows the strongest influence on curiosity among the four latent functions (2.612 $ \pm $ 0.124). The natural merging of knowledge identification and knowledge acquisition corroborates with the notation that one person's knowledge-seeking may draw the attention of another group member to a related knowledge gap and escalate collaborative knowledge-seeking. Behaviors that positively contribute to this function are mainly from cluster 3 ({\em sharing findings, task-related question asking, argument, and evaluation of other's ideas}). Also, nonverbal behaviors including {\em head turn} and {\em turn taking dynamics (indegree)} are also related to this function, which support the idea that a higher degree of group members' interest and involvement in the social interaction stimulates awareness of peer's ideas, subsequently leading to knowledge-seeking via social means to gain knowledge from the experience of others, and add that onto one's own direct experiences.

Second, {\em Individual Knowledge Identification and Acquisition} shows a strong influence on curiosity (2.149 $\pm$ 0.066). Similar to the interpersonal level function, knowledge identification and acquisition merge into one coherent function, as knowledge-seeking behaviors can sparkle new unknown or conflicting information within the same individual. Behaviors from cluster 2 ({\em hypothesis generation, justification, idea verbalization}) and cluster 4 ({\em confusion, joy, surprise, uncertain, positive sentiment towards task}) mainly contribute to this function. {\em Head nod}, indicative of positive feelings towards the stimulus due to its compatibility with the response \cite{forster1996influence}, maps to this function as well. Finally, {\em turn taking (indegree and outdegree)} and {\em social question asking} contribute positively to individual knowledge identification and acquisition. Interest in others reflects a general level of trait curiosity \cite{renner2006curiosity}.

Third, we find that a relatively small group of behaviors including {\em agreement, idea verbalization} and {\em lateral head inclination} have a predictive influence on the latent function of {\em Interpersonal Knowledge Intensification}, which in turn has a high positive influence on curiosity (1.756 $\pm$ 0.238). Agreement may contribute to information seeking by promoting acceptance and cohesion. Working in social contexts broadcasts idea verbalization done by an individual to other group members, which might in turn increase their willingness to get involved. Lateral head inclination during the RGM activity is associated with an intensive investigation of the RGM solution offered by both oneself and other group members. Overall, engagement in a cooperative effort to overcome common blocking points in the group work may result in intensifying knowledge-seeking.

Finally, the latent function of {\em Individual Knowledge Intensification} has the least comparative influence on curiosity. It is associated with nonverbal behaviors such as {\em head nod} and emotional expressions of positive affect ({\em flow, joy} and {\em delight}), which function towards increasing pleasurable arousal. Also, {\em surprise} and {\em suggestion} positively influence this latent function and signal increased anticipation to discover novelty, conceptual conflict, and correctness of one's idea. Interestingly, results also show that {\em negative sentiment about the task} positively influences an individual's knowledge-seeking behaviors. Qualitative corpus observations reveal that such verbal expressions often co-occur with the evaluation made by a group member within the same 10-second thin-slice, signaling a desire for cooperation. Thus, a potential explanation of this association is that expressing negative sentiment about the task may signal hardship, which draws group member's attention and increases chances of receiving assistance, thus increasing engagement in knowledge-seeking.
\vspace{-0.7cm}

\subsection{Causal Intra/Inter-personal Influence Patterns}
\vspace{-0.1cm}
The CTSEM analyses described above establish the overall degree of predictive influence among verbal/nonverbal behaviors, latent functions and ground truth thin-slice curiosity. However, they do not provide ``causal" insights into inter-and-intra-personal curiosity dynamics of the kind that can be leveraged by a learning technology to foster curiosity in group work. We therefore conducted follow-up causal analyses with the dual goal of generating such actionable insights and providing additional empirical evidence for our proposed theoretical framework of curiosity.

Specifically, we ran conditional Granger causality \cite{ding2006granger} to assess the interdependence among verbal/nonverbal behaviors and their causal contribution to the increase (0$\rightarrow$1, 0$\rightarrow$2, 1$\rightarrow$2) or maintenance (1$\rightarrow$1, 2$\rightarrow$2) of the thin-slice curiosity level (or vice-versa). This notion of causal influence is based on the idea that if the variance of the autoregressive prediction error of time-series $A$ at the present time is reduced by inclusion of past measurements from time-series $B$, then $B$ is said to have a causal influence on $A$. Because such a causal relation (based on cause-effect relations with constant conjunctions) between $A$ and $B$ can be direct, mediated by a third time-series $C$, or be a combination of both, the technique of conditional Granger causality allows modeling causal relationship among multivariate behavioral time series. More technical details can be found in \cite{sinhaectel2}.

Here we report four categories of causal influence patterns in Table 4. First, the top-left corner highlights the causal influence of other's past behavior (within the last minute) on one's increase/maintenance of curiosity. Such triggering behaviors span both verbal (e.g., justification, idea verbalization, task question asking, negative evaluation, etc) and nonverbal (e.g., head movement variability, confusion and surprise-related facial expressions) categories. Second, the top-right corner highlights the reciprocal pattern of causal influence, i.e., other's increase/maintenance of curiosity in the past (within the last minute) on one's behaviors. Such triggered behaviors include not just uncritical acceptance in the form of a head nod or verbal agreement, but also those that signal investment into furthering knowledge acquisition like on-task question tasking, justification and flow-related facial expressions. Taken together, these bi-directional sets of interpersonal causal influence patterns provide concrete data-driven behaviors for a virtual peer (or a different learning technology) to emulate for causing an increase/maintenance in curiosity ($+\Delta$) for a group member. The bottom-left and bottom-right corners of Table 4 highlight a comparatively small(er) number of intrapersonal causal influence patterns, those where one's past behavior (within the last minute) leads to an increase/maintenance of curiosity (and vice-versa).

\begin{table*}[ht]
\caption{Conditional Granger causality results (at 1\% level of significance, 1 minute granularity) between verbal/nonverbal behavior time series and thin-slice curiosity (increase/maintenance) time series. Interpersonal causal influence patterns reflect interdependence of {\em other's} behavior on one's own curiosity $+\Delta$ (top left), or {\em other's} curiosity $+\Delta$ on one's own behavior (top right). Intrapersonal causal influence patterns reflect interdependence of {\em one's own} behavior on curiosity $+\Delta$ (bottom left), or {\em one's own} curiosity $+\Delta$ on behavior (bottom right).}
\vspace{-0.1cm}
\resizebox{\textwidth}{!}{%
\begin{tabular}{c | ccc | cccc}
 & \textbf{\begin{tabular}[c]{@{}c@{}}Causal Influence\\ (From)\end{tabular}} & \textbf{\begin{tabular}[c]{@{}c@{}}Causal Influence\\ (To)\end{tabular}} & \textbf{\begin{tabular}[c]{@{}c@{}}Standardized Causal \\Strength (G-ratio)\end{tabular}} & \multirow{13}{*}{} & \textbf{\begin{tabular}[c]{@{}c@{}}Causal Influence\\ (From)\end{tabular}} & \textbf{\begin{tabular}[c]{@{}c@{}}Causal Influence\\ (To)\end{tabular}} & \textbf{\begin{tabular}[c]{@{}c@{}}Standardized Causal \\Strength (G-ratio)\end{tabular}} \\
\multirow{12}{*}{\textbf{Interpersonal}} & Confusion-related facial expressions & \multirow{12}{*}{Curiosity$_{increase/maintain}$} & 0.591 &  & \multirow{12}{*}{Curiosity$_{increase/maintain}$} & Question Asking (on-task) & 0.834 \\
 & Surprise-related facial expressions &  & 0.420 &  &  & Head Nod & 0.648 \\
 & Head Turn &  & 0.303 &  &  & Justification & 0.281 \\
 & Head Nod &  & 0.222 &  &  & Turn Taking Outdegree & 0.276 \\
 & Justification &  & 0.208 &  &  & Joy-related facial expressions & 0.225 \\
 & Task Sentiment (positive) &  & 0.198 &  &  & Agreement & 0.138 \\
 & Idea Verbalization &  & 0.196 &  &  & Flow-related facial expressions & 0.057 \\
 & Joy-related facial expressions &  & 0.159 &  &  & Lateral Head Inclination & 0.001 \\
 & Evaluation (negative) &  & 0.151 &  &  &  &  \\
 & Lateral Head Inclination &  & 0.083 &  &  &  &  \\
 & Question Asking (on-task) &  & 0.046 &  &  &  &  \\
 & Turn Taking Outdegree &  & 0.025 &  &  &  &  \\
\multicolumn{4}{c}{} &  & \multicolumn{3}{c}{} \\
\multirow{5}{*}{\textbf{Intrapersonal}} & Uncertainty & \multirow{5}{*}{Curiosity$_{increase/maintain}$} & 1 & \multirow{5}{*}{} & \multirow{5}{*}{Curiosity$_{increase/maintain}$} & Confusion-related facial expressions & 0.306 \\
 & Lateral Head Inclination &  & 0.863 &  &  & Uncertainty & 0.278 \\
 & Joy-related facial expressions &  & 0.520 &  &  & Task Sentiment (negative) & 0.238 \\
 & Head Nod &  & 0.134 &  &  & Head Nod & 0.2 \\
 & Task Sentiment (negative) &  & 0.021 &  &  & Task Sentiment (positive) & 0.026
\end{tabular}%
\vspace{-0.2cm}
}
\end{table*}

\vspace{-0.35cm}
\subsection{Implementation in Pedagogical Agents}
\vspace{-0.1cm}

\begin{figure}[h]
\centering
\vspace{-0.1cm}
\includegraphics[width=5.9cm]{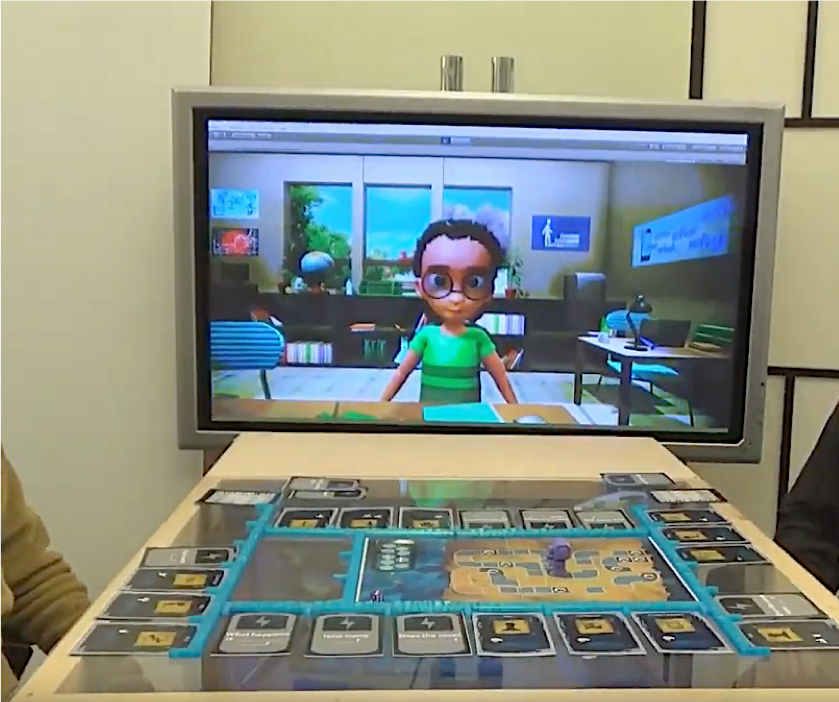}
\vspace{-0.3cm}
\caption{Virtual peer interacts with the customized Outbreak board game.}
\vspace{-0.1cm}

\end{figure}

We have developed a virtual peer that aims to elicit curiosity for young children in a collaborative tabletop game named Outbreak \cite{to2018modeling}. We applied heuristics derived from our theoretical framework of curiosity as described in this article and in \cite{sinhaectel1}, as well as our subsequent computational model of curiosity \cite{sinhaectel2,paranjape2018predicting}. The objective was to design social interactions for the virtual peer to perform for evoking key curiosity drives which lead to identification and acquisition of knowledge. First, we developed a virtual peer whose appearance resembles a 9-14 years old child with ambiguous gender and ethnicity (to avoid inadvertent gender and ethnicity stereotypes that might lead to stereotype threat for particular groups of students \cite{shapiro2012role}). Figure 5 illustrates the virtual peer setup during a gameplay session.

The architecture underlying the virtual peer resembles a dialogue system architecture. The first module developed has been a verbal behavior generation module \cite{Paranjape2018Towards} that allows the virtual peer to produce age-appropriate curiosity-inducing conversational moves including question asking, hypothesis verbalization, argument, and justification during the discussions elicited by the collaborative board game. This generation module allows the virtual peer to generate the kind of sentences that the human-human analyses described above have shown to increase curiosity.

\vspace{-0.3cm}
\section{Implications}
\vspace{-0.1cm}
Few current learning technologies aim to foster the metacognitive factors shown to increase learning (although see \cite{conati2000toward,aleven2006toward,azevedo2009metatutor,pifarre2010promoting,bannert2012supporting}). We believe reasons for this include (i) lack of theoretical formalisms and real-time measurement approaches to capture the intricate nature of metacognitive and socio-emotional factors such as creativity, curiosity, grit, help-seeking, self-explanation ability, etc \cite{duckworth2015measurement}, and (ii) dearth of an operational way to embed this theoretical understanding into computational models that can leverage mapping between behaviors and their underlying mechanisms to offer scaffolding \cite{reiter2017theories}. Here, we attempted to go beyond former research on curiosity inference directly from visual and vocal cues (e.g., \cite{craig2008emote,baranes2015eye,nojavanasghari2016emoreact}), by considering underlying mechanisms that link these low-level cues to curiosity, and empirically validating how such cues interact with group collaboration, interpersonal dynamics and conversation moves. Our research approach holds concrete implications for the design of learning technologies, and multimodal learning analytics.

\vspace{-0.35cm}
\subsection{Implications for Learning Technologies}
\vspace{-0.1cm}
For learning technology design, our prior work has demonstrated that one successful approach is to investigate what forms of verbal and nonverbal behaviors and their corresponding functions are good indicators of curiosity - and good elicitors of curiosity - in human-human interaction. This can facilitate the development of adaptive technologies that look for opportunities to use strategies (that is, to be tactical) to scaffold, maintain and evoke curiosity. The underlying infrastructure needed to translate the proposed theoretical framework into teaching tools and technological learning environments should comprise (i) a curiosity perception module for accurate multimodal behavior sensing using available sensors (e.g., cameras, microphones and other biometric devices) in real-time interaction, (ii) a curiosity reasoner that outlines how a learning technology should act in to support those same functions, and a generation module to translate the intentions generated by the reasoner into observable verbal and nonverbal behaviors for the learning technology to perform.

Prior work on pedagogical agents/robots \hspace{-0.008cm}\cite{wu2013innovative} makes assumptions regarding what observable behaviors can represent lack of curiosity and should therefore result in a move on the part of the learning technology. Here, instead, we relied on close observation of human interaction to extract the moment-to-moment behaviors that occur during periods of curiosity, and amplified the power of those correlations by introducing the theoretically grounded layer of functions that can be fulfilled by behaviors. Below, we expand on how these functions can be scaffolded using strategies and associated tactics in virtual peers, in other technologies, and even simply by teachers or tutors. Specific strategies and tactics are needed because a virtual peer acting curious is not sufficient. Prior literature \cite{morrisonmoves} and our empirical data show that (i) a curious child may not always attempt to increase the curiosity of another child (nor does curiosity in one child always cause curiosity in another) perhaps because it requires increased cognitive and social effort, and (ii) disinterested children may become increasingly cut-off from the core interactions of the group over time, and therefore it's important to find ways to reignite their interest and engagement. In what follows we summarize the strategies and tactics we found in group informal learning contexts. These can be implemented in a virtual peer, and other learning technologies.

\subsubsection{Strategies for Supporting Functions of Curiosity}
We define strategies as moves made by a third party (learning technology or human coach/peer) in the service of facilitating curiosity. This means that strategies serve as vehicles for an influence attempt in the group, directly (e.g., affecting gains/costs) or indirectly (e.g., controlling critical environmental aspects) affecting the behavior of a group member \cite{cartwright1953group}. We believe that strategies should support underlying functions that contribute to curiosity because this enables computer support to address the root cause of undesirable behavior or provide a reinforcing means for the root cause of desirable behavior. The success of a strategy can be determined concretely by the extent to which it increases thin-slice curiosity in subsequent time-intervals, and specifically by the probability that it leads to the expected behavior(s) by a target child. We propose a categorization of strategies into three clusters in alignment with the functions of curiosity.

Strategies that can facilitate the {\em knowledge identification} function should help peers realize potential new knowledge to seek. Our empirical data suggested awareness of novel or complex stimuli as a means towards this end. Prior literature also highlights that raising awareness of the conflict between group members can facilitate knowledge identification \hspace{-0.008cm}\cite{forsyth2009group}. Strategies that can facilitate the {\em knowledge acquisition} function should stimulate critical thinking by helping peers develop new interpretations and consider alternative perspectives, and prevent them from valuing their cohesiveness and relationships with others so much that they avoid conflict and challenging each other's ideas \cite{cartwright1953group}. Literature suggests that provoking group members out of their comfort zone and encouraging them to rethink/defend their responses can be a means towards this end \cite{correnti2015improving}. Strategies that can facilitate the {\em intensification of knowledge identification and acquisition} function may create a friendly climate, and honor the seeking of knowledge gained through trial and error. Our empirical data in formal learning contexts suggested that when learners were explicitly told to avoid personal criticism and pro-actively grapple with their intuitive ideas, they often opened up to their peers, with a reduced fear of being seen as incompetent, or being excluded. Literature also emphasizes the provision of a supportive and psychologically safe environment to have beneficial effects on accelerated risk-taking \cite{edmondson1999psychological}.

\subsubsection{Tactics for Exercising a Particular Strategy}
We define tactics as particular observable ways in which a strategy can be exercised. The rationale behind the usage of tactics is that certain forms of group interaction (e.g., sharing findings with peers) are more effective for raising curiosity than others. Thus, it is worthwhile to make some arrangements for provoking such interactions, as they may not occur spontaneously. This approach is inspired by the traditional computer-supported collaborative learning literature on scripting \cite{fischer2007scripting}, however, we propose a more real-time version of such scripting-based approaches. Our interest is rather in adaptive regulation of the small group interaction ``on the fly" \cite{jermann2008group} by continuously comparing the current curiosity level with a target configuration (e.g., if the likelihood of being in high curiosity level in the following time intervals exceeds a certain threshold), and exercising tactics to restore equilibrium whenever there is a discrepancy in the current and target curiosity level. We propose a categorization of tactics falling into each of the three strategy clusters.

Our observations show scarce evidence of children explicitly making moves to exercise the strategy of {\em knowledge identification}. However, the literature suggests that attentional anchors in the form of contrasting case comparisons can be leveraged by a coach towards this end \cite{alfieri2013learning}. By creating paradoxes, such a tactic can help children notice novel features of the stimulus. Our data do bring to light different tactics children use in the service of exercising the strategy of {\em knowledge acquisition}. These tactics include challenging a peer's responses (e.g., I don't think this will be really sturdy though), and asking them to make an explicit link between ideas, representations and solution strategies (e.g., what's your evidence for that?). Prior literature also highlights how tactics such as (i) making group members take positions on a big question raised by a task issue and then present reasons and evidence for and against, (ii) encouraging group members who are in conflict to paraphrase each other's position, can be used in service of similar goals \cite{correnti2015improving}. Finally, we observe that tactics children use for exercising the strategy of {\em intensifying knowledge identification or acquisition} include expressing curiosity-orientation behavior (e.g., looking at stimulus with surprise, expressing interest in individuals and activities using gaze or body orientation), and excitement about solution strategy of a peer. Concrete tactics such as rewarding risk-taking and providing positive feedback for effort may have a positive impact on fulfilling these goals.
\vspace{-0.33cm}

\subsection{Implications for Multimodal Learning Analytics}
\vspace{-0.1cm}
Joining a long tradition of multimodal human behavior analysis in the Learning Sciences (e.g., \cite{ryokai2002literacy,koschmann2002learner}), and other application areas, Multimodal Learning Analytics (MMLA) \cite{blikstein2013multimodal} offers a useful lens through which to interpret the sensing of, reasoning about and responding to natural human behavior in learning environments, including both verbal and nonverbal devices. In the current work, we developed a theoretical model of curiosity, built arguments and adduced evidence for multimodal indicators and elicitors of curiosity, combined machine learning and human annotation to refine the theoretical model, and ended with a statistical model that tests and validates the theoretical assumptions.

While we have primarily focused on the implications of this work for the implementation of virtual peers, another straightforward implication of our work bears on the tension in MMLA between preserving enough of the complexity of curiosity-driven behavior in the theoretical framework (to obtain valid scientific insight), while at the same time attaining computational feasibility (making sure annotated behavior representations retain enough contextual information and are learnable by machine learning based algorithmic approaches). The part of our analyses related to sensing highlights a potential multimodal fusion solution of capturing and synchronizing different streams of data at the grain size of the ground truth (in this instance thin-slice annotation of curiosity).

Also relevant to MMLA is our iterative theory-and-data-driven approach for capturing not just noisy and intermittent low-level data gathered by physical sensors, but also translating such data into models that describe the underlying psychological states that are correlated with and sometimes cause those observable behaviors. These mid-level phenomena are more generalizable across learning contexts and across metacognitive phenomena. In fact, little work in MMLA has addressed the temporal dimensions of relationships among verbal and nonverbal behaviors, and CTSEM may provide a useful tool in looking at how \textit{series} of behaviors may predict an underlying state such as curiosity.

These MMLA implications bring with them practical challenges that include (i) combating inaccuracy in detecting behavioral episodes because machine learning approaches are not 100\% robust, (ii) developing consistent and opportunistic planning algorithms for selecting responding strategies based on current percepts and tying together associated sets of tactics because human behavior is not always rational and dynamic environmental circumstances may affect pre-planned action recipes, (iii) using appropriate evaluation criteria to test the effectiveness of a responding strategy on subsequent behavior in human-technology interaction because people's application of social rules and heuristics from their domain to the domain of machines is often moderated by socially-competent behavior exhibited by the machine \cite{cassell2007intersubjectivity}.

\vspace{-0.3cm}
\section{Limitations}
\vspace{-0.12cm}
We must acknowledge some methodological limitations of this work. First, the small sample size limits generalizability. Second, the reliability of thin-slice annotation of ground truth (curiosity) via crowdsourcing platforms can be improved by varying more carefully factors such as the time-scale (granularity) of ratings, rating scale, choice of crowdsourcing workers, and task setup on the crowdsourcing platform. Third, our approach of combining machine annotation with human judgment for annotation of verbal behaviors increases reproducibility, speed and scalability, without compromising on inter-rater reliability. Despite these advantages, going through machine annotated labels and evaluating their accuracy (de-noising) is a different task than if those labels were not there in the first place (meaning that a completely manual annotation approach had been followed). Future work could have some intermediate points during this de-noising process, where the initial inter-rater reliability for human judgment (left column of Table 3) could be re-evaluated for consistency. Fourth, latent variable models used as empirical validation tools in our work are limited by their ability to make causal inferences, especially in cross-sectional datasets \cite{nagengast2015prospects}.

Fifth, we used a crude proxy to infer emotional states from facial landmarks, and future work could adopt complementary predictive modeling approaches \cite{sariyanidi2015automatic}. Although facial expressions have the advantage of being observable and recognizable using current computer vision approaches with high accuracy, they can often be polysemous, ambiguous, and be voluntarily camouflaged for social reasons. That is, a smile may mean embarrassment and not happiness, it's unclear where a head nod means agreement or simply that one follows what the other person is saying, and cultural differences include covering facial expressions with one's hand because it is inappropriate to smile in a company (for example). Such subtle distinctions among underlying mechanisms cannot be teased apart. Sixth, it is important to note that despite the existence of learning opportunities, a given student's unwillingness to learn and explore is over-determined. It can stem from multiple sources, such as unawareness of new information that is to be learned, lack of knowledge (competency) and information-seeking skills, lack of environmental support, shyness in talking to peers, and so forth. Therefore, although our research sketched a detailed outline of individual and interpersonal functions of curiosity, this is only a first step. The next phase of this research will use the virtual peer as a way of assessing the validity of our model by observing the impact of given functions and given behaviors on the human interlocutors.


\vspace{-0.3cm}
\section{Future Work}
\vspace{-0.1cm}
To examine the effectiveness of the computational model in eliciting curiosity during agent-child interaction, we are developing a semi-automated Wizard-Of-Oz system that integrates (i) the curiosity elicitation reasoner to decide specific real-time contexts during which verbal and non-verbal responses of the virtual peer can elicit curiosity in group work, (ii) the behavior understanding module to understand real-time game and behavior updates of multiple game players. This kind of assessment of the model allows us to turn behaviors on and off to see which ones have the highest impact on curiosity (a luxury not afforded by observing child-child interaction).

We have also begun to extend our analysis of nonverbal behaviors beyond facial expressions to other parts of the body. We are therefore beginning to analyze the nature of hand movements to understand their role in indexing and in triggering curiosity. Hand movements, including manipulative actions and communicative gestures \cite{pavlovic1997visual}, are a key aspect of hands-on collaborative learning. From a distributed cognition perspective, manipulative actions on artifacts and external representations embedded in the physical surroundings help shape people's thoughts \cite{hutchins2006distributed} by serving important epistemic functions in information gathering and ease of cognitive work \cite{kirsh1994distinguishing}, and supporting communication \cite{cassell1999speech}.
Although hand movements have long been recognized as indispensable to the kinds of physical and epistemic exploration that contribute to curiosity \cite{berlyne1960conflict}, there is a research gap in identifying quantitative relationships between hand movement and curiosity through empirical observation. As an extension of the theoretical framework of curiosity outlined here, we will conduct analyses of key relationships \cite{fournier2012mining} between hand movements and verbal behaviors. As inputs into these analyses, we have articulated a taxonomy of such hand movements with respect to form and meaning. Specifically, we have narrowed down gestures to three kinds of hand movements commonly found in our in-lab study corpus, based on McNeill's well-known gesture coding scheme \cite{mcneill1992hand}.

\vspace{-0.35cm}
\section{Conclusion}
\vspace{-0.1cm}
The results, model, and framework presented here are part of a larger research effort to understand the social scaffolding of curiosity and to use that understanding to implement a virtual peer to increase children's curiosity in formal and informal learning contexts, and from there their ability and desire to learn. The theoretical framework presented in this article lays the foundation of a computational model of curiosity that can enable a virtual peer to sense the real-time curiosity level of each member in small group interaction, and the impact of each member's behavior on the others. Here, we articulated key social factors that account for curiosity in learning in social contexts, proposed and empirically validated a novel theoretical framework that disentangles individual and interpersonal functions linked to curiosity and behaviors that fulfill these functions.

The framework connects observable behaviors of children with their underlying curiosity state. Our analyses reveal interpersonal dynamics that indicate as well as elicit curiosity states. As indicators of a curiosity state, the framework reveals strong positive predictive relationships between the interpersonal functions of knowledge identification, acquisition and intensification on curiosity, whereby individual peers had an impact on one another. In terms of eliciting curiosity states, the framework identifies interpersonal behavior patterns that causally influence the increase or maintenance of curiosity. Taken together, these results provide strong evidence for the social nature of curiosity, and the need to disentangle its interpersonal precursors from its individual precursors. In the context of a science game, a pedagogical agent that speaks like a child was also developed to generate the kind of sentences that our current analyses of human-human data have shown to result in increased curiosity. Through designing such learning technologies, we hope to provide additional pedagogical supports to help children develop knowledge-seeking skills driven by intrinsic curiosity.

\vspace{-0.25cm}
\section*{Acknowledgment}
\vspace{-0.1cm}
We would like to thank the Heinz Endowment for their generous support, all student interns who worked with us, teachers and instructors of local schools and summer camps that so kindly allowed us to observe, as well our collaborators Dr. Jessica Hammer, Dr. Louis-Philippe Morency, Dr. Geoff Kaufman, Alexandra To and Bhargavi Paranjape for supporting the SCIPR project.
\vspace{-0.25cm}

\ifCLASSOPTIONcaptionsoff
  \newpage
\fi



\bibliographystyle{IEEEtran}
\bibliography{splncs}
%

\vspace{-9cm}

\begin{IEEEbiographynophoto}{Tanmay Sinha}
 is a postdoctoral researcher in Learning Sciences and Technology Design at ETH Zurich. His doctoral work focused on studying the pedagogical value of deliberate, guided failure to foster students' conceptual understanding and transfer. He is interested in the role of socio-emotional and interpersonal factors in learning, and more generally, in how scaffolded human and technology interventions may open new opportunities for students to fully reap the benefits of failure-driven exploration.
\end{IEEEbiographynophoto}

\vspace{-10cm}

\begin{IEEEbiographynophoto}{Zhen Bai} is an Assistant Professor in the Department of Computer Science at University of Rochester. She is interested in designing embodied and intelligent interfaces that support lifelong learning and quality of life for people with diverse abilities and backgrounds by augmenting social interaction and meaning making in a playful and collaborative manner.
\end{IEEEbiographynophoto}

\vspace{-10cm}

\begin{IEEEbiographynophoto}{Justine Cassell}
is SCS Dean's Professor in the School of Computer Science at Carnegie Mellon University, and until recently was Director of the Human-Computer Interaction Institute in the School of Computer Science. She is currently on leave from Carnegie Mellon University to hold the founding international chair at PRAIRIE Paris Institute on Interdisciplinary Research in AI (one of French President Macron's four new AI Institutes), and hold the associated position of Directrice de Recherche at Inria Paris. Her research focuses on understanding natural forms of communication, and then creating technological tools for those forms of communication and linguistic expression to flourish in the digital world.
\end{IEEEbiographynophoto}

%




\end{document}